%% file: iros2019_PlaceCat.tex
\documentclass[letterpaper, 10 pt, conference]{ieeeconf} 

\IEEEoverridecommandlockouts                              

\usepackage{amsmath, amssymb , graphicx}
\usepackage{multirow}
\usepackage{multicol}
\usepackage{siunitx}
\usepackage{color}
\usepackage{array}
\usepackage{makecell}
\usepackage{hyperref}
\hypersetup{colorlinks=true}
\usepackage{xargs}
\newcommand{\etal}{\textit{et al}.}
\usepackage{dblfloatfix} 
\usepackage[colorinlistoftodos,prependcaption,textsize=tiny]{todonotes}
\newcommandx{\unsure}[2][1=]{\todo[linecolor=red,backgroundcolor=red!25,bordercolor=red,#1]{#2}}

\usepackage{hyperref}
\hypersetup{
    colorlinks=true,
    linkcolor=blue,
    filecolor=blue,      
    urlcolor=blue,
}

\usepackage{subcaption}
\usepackage [english]{babel}
\usepackage [autostyle, english = american]{csquotes}
\MakeOuterQuote{"}

\let\svfootnoterule\footnoterule
\renewcommand\footnoterule{\vfill\svfootnoterule}

\title{\LARGE \bf DEDUCE: Diverse scEne Detection methods in Unseen Challenging Environments}

\author{Anwesan Pal \and Carlos Nieto-Granda \and Henrik I. Christensen \thanks{Contextual Robotics Institute, University of California, San Diego.}
\thanks{La Jolla, CA 92093, USA.} \thanks{\tt\footnotesize \{a2pal, cnietogr,  hichristensen\}@eng.ucsd.edu}}

\begin{document} 

\bstctlcite{IEEEexample:BSTcontrol}

\maketitle
\thispagestyle{empty}
\pagestyle{empty}

\begin{abstract}
In recent years, there has been a rapid increase in the number of service robots deployed for aiding people in their daily activities. Unfortunately, most of these robots require human input for training in order to do tasks in indoor environments. Successful domestic navigation often requires access to semantic information about the environment, which can be learned without human guidance. In this paper, we propose a set of DEDUCE\footnote{Supplementary material including code and the videos of the different experiments are available at~\url{https://sites.google.com/eng.ucsd.edu/deduce}.} - \textit{Diverse scEne Detection methods in Unseen Challenging Environments} algorithms which incorporate deep fusion models derived from scene recognition systems and object detectors. The five methods described here have been evaluated on several popular recent image datasets, as well as real-world videos acquired through multiple mobile platforms. The final results show an improvement over the existing state-of-the-art visual place recognition systems.
\end{abstract}

%


\input{files/introduction.tex}
\input{files/relatedwork.tex}
\input{files/methodology.tex}
\input{files/expresults.tex}
\input{files/conclusion.tex}
\input{files/acknowledgements.tex}
\bibliographystyle{IEEEtran}
\bibliography{references}

\input{files/appendix.tex}

\end{document}

%% file: files/introduction.tex
\section{Introduction}\label{sec:intro}
Scene recognition and understanding has been an important area of research in the robotics and computer vision community for more than a decade now. Programming robots to identify their surroundings is integral to building autonomous systems for aiding humans in house-hold environments.

Kostavelis \etal~\cite{Kostavelis2015:RAS} have recently provided a survey of previous work in semantic mapping using robots in the last decade. According to their study, scene annotation augments topological maps based on human input or visual information of the environment. Bormann \etal~\cite{Bormann2016:ICRA} pointed out that the most popular approaches in room segmentation involve segmenting floor plans based on spatial regions.  

An essential aspect of any spatial region is the presence of specific objects in it. Some examples include a bed in a bedroom, a stove in a kitchen, a sofa in a living room, etc. S\"{a}nderhauf \etal~\cite{Niko2018:IJRR} formulated the following three reasoning challenges that address the semantics  and  geometry  of a scene and the objects therein, both separately and jointly:  1) Reasoning  About  Object  and  Scene  Semantics, 2)  Reasoning  About  Object  and  Scene  Geometry, and 3) Joint Reasoning about Semantics and Geometry. This paper focuses on the first reasoning challenge and uses Convolutional Neural Networks (CNNs) as feature extractors for both scenes and objects. Our goal is to design a system that allows a robot to identify the area where it is located using visual information in the same way as a human being would.  

\textbf{Contribution}. We consider five models of scene prediction through the integration of object and scene information in order to perform place categorization. We then evaluate the robustness of these models by conducting extensive experiments on state-of-the-art still-image datasets, real-world videos captured via various hand-held cameras, and also those recorded using a mobile robot platform in multiple challenging environments. One such environment is shown in Figure \ref{fig:fetch_semMapping}, where the segmented regions correspond to our scene detection. The results obtained from the experiments demonstrate that our proposed system can be generalized well beyond the training data, and is also impervious to object clutter, motion blur, and varying light conditions. 

The paper is organized as follows. Section~\ref{sec:relwork} provides a detailed list of recent works in place categorization. The proposed DEDUCE algorithms are described in details in Section~\ref{sec:method}. Section ~\ref{sec:exp_res} summarizes the diverse experiments conducted and the results that were obtained. Finally, Section~\ref{sec:conclusion} concludes the paper and discusses the future work towards real-world domestic, search and rescue tasks with robots in indoor environments. 
 
\input{images/fetch_semMapping.tex}

%% file: images/fetch_semMapping.tex
\begin{figure}[!t]
\centering
    \includegraphics[width=0.48\textwidth]{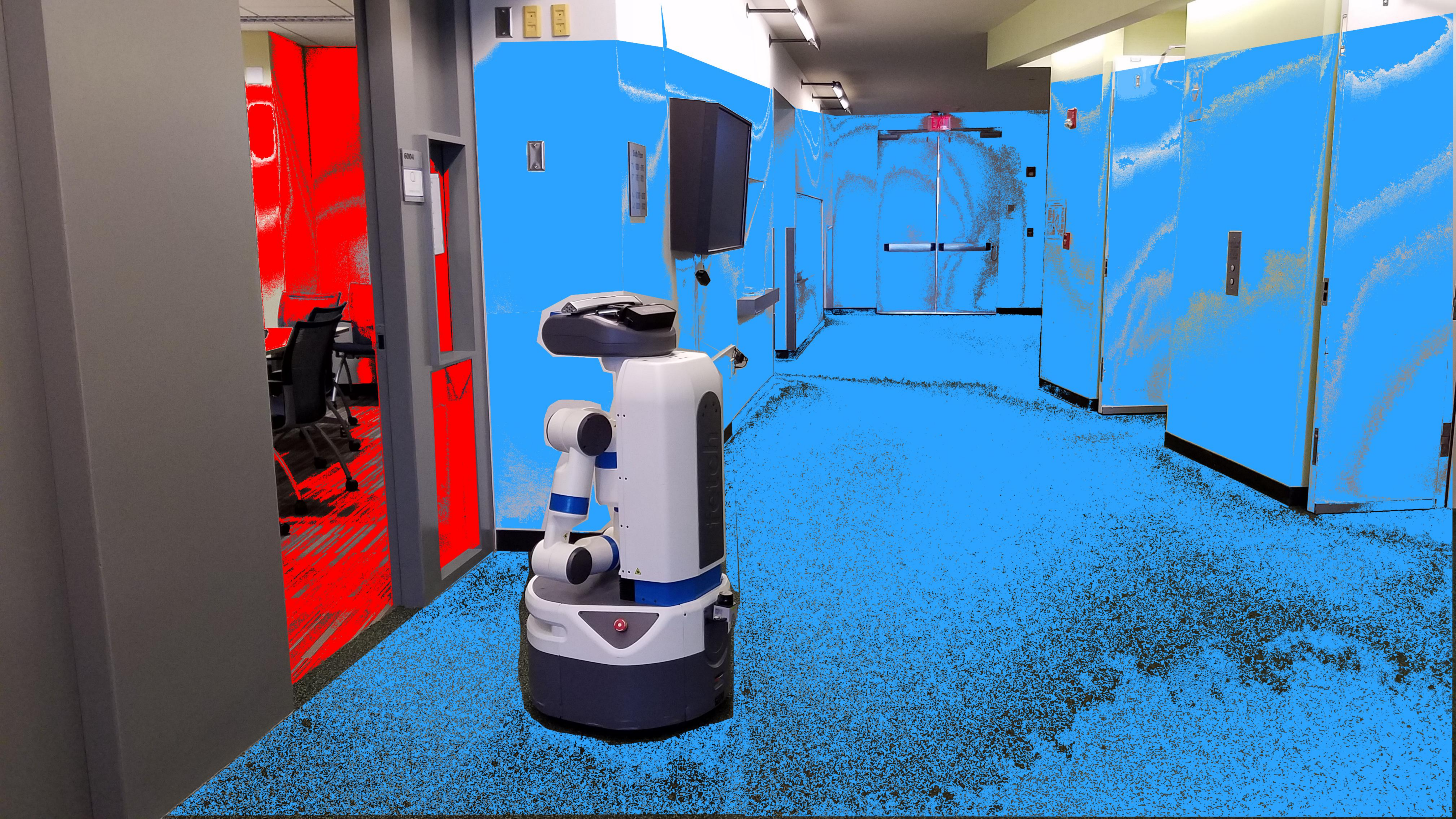}
    \caption{\small Visualization of the semantic mapping performed while the Fetch robot is navigating through the environment (best viewed in color). The associated map is shown in Figure \ref{maps:atkinson}.}
    \label{fig:fetch_semMapping}
\end{figure}

%% file: files/relatedwork.tex
\section{Related Work}\label{sec:relwork}
Semantic place categorization using only visual features has been an important area of research for robotic applications~\cite{Torralba2003:ICCV, Wu2009:IROS}. In the past, many robotics researchers focused on place recognition tasks~\cite{Pronobis2008:ICRA, Siagian2007:PAMI} or on the problem of scene recognition in computer vision~\cite{oliva2001:IJCV, Quattoni2009:CVPR}.

Quattoni and Torralba~\cite{Quattoni2009:CVPR} introduced a purely vision-based place recognition system that improved the performance of the global gist descriptor by detecting prototypical scene regions. However, the annotations regarding the size, shape and location of up to ten object prototypes must be provided, and learned in advance for their system to work. Also, the labeling task is very work-intensive, and the approach of having fixed regions is only useful in finding objects in typical views of the scene. This makes the system ill-suited for robotics applications. Since we want to deal with flexible positions of objects, we apply a visual attention mechanism that can locate important regions in a scene automatically.

A number of different approaches have been proposed to address the problem of classifying environments. One popular approach adopted is to use feature matching with Simultaneous Localization and Mapping (SLAM). Ekvall \etal~\cite{Ekvall2007:Robotica} and Tong \etal~\cite{tong2017sceneslam} demonstrated a strategy for integrating spatial and semantic knowledge in a service environment using SLAM and object detection \& recognition based on Receptive Co-occurrence Histograms. Espinace \etal~\cite{Espinace2010:ICRA} presented an indoor scene recognition system based on a generative probabilistic hierarchical model using the contextual relations to associate objects to scenes. Kollar \etal~\cite{Kollar2009:ICRA} utilized the notion of object to object and object to scene context to reason about the geometric structure of the environment in order to predict the location of the objects. The performance of the object classifiers is improved by including geometrical information obtained from a 3D range sensor. In addition, it also facilitates a focus of attention mechanism. However, these approaches only identify the place based on the specific objects detected and the hierarchical model that is used to link the objects with the place. In contrast to their method, our algorithm is not limited to a small number of recognizable objects, since we also take scene semantics into consideration.

Recently, there have been several approaches to scene recognition using artificial neural networks. Liao \etal~\cite{liao2016understand, liao2017place} used CNNs for recognizing the environment by regularizing deep architecture with semantic segmentation through the occurrence of objects. However, their system information and mapping results are not provided, and they also do not conduct any cross-domain analysis. Sun \etal~\cite{sun2018scene} proposed a unified CNN which performs scene recognition and object detection as a multi-task problem. Luo \etal ~\cite{Luo2018HierarchicalSM} developed a semantic mapping framework utilizing spatial room segmentation, CNNs trained for object recognition, and a hybrid map provided by a customized service robot. S\"{a}nderhauf \etal~\cite{sunderhauf2016place} proposed a transferable  and  expandable place categorization and semantic mapping system that requires  no  environment-specific  training. Mancini \etal~\cite{mancini2018robust} addressed Domain Generalization (DG) in the context of semantic place categorization. They also provide results of state-of-the-art algorithms on the VPC Dataset~\cite{Wu2009:IROS} that we compare our performance with. However, most of these results do not test their algorithms on a wide variety of platforms like we do in this paper, both for static images, and dynamic real world videos captured using different hand-held cameras and mobile robotic platforms.

%% file: files/methodology.tex
\section{Methodology} \label{sec:method}
We consider a set of five different models, abbreviated as Diverse scEne Detection methods in Unseen Challenging Environments (DEDUCE), for place categorization. Each model is derived from two base modules, one based on the PlacesCNN~\cite{zhou2017places} and the other being an Object Detector-YOLOv3~\cite{redmon2018:arxiv}. The classification model can be formulated as a supervised learning problem. Given a set of labelled training data $\mathcal{X}^{tr} = \{(\textbf{x}_1,\textbf{y}_1),(\textbf{x}_2,\textbf{y}_2)...(\textbf{x}_N,\textbf{y}_N)\}$, where $\textbf{x}_i$ corresponds to the data samples and $\textbf{y}_i$ to the scene labels, the classifier should learn the discriminative probability model
\begin{equation}
    p(\hat{\textbf{y}}_j|\Phi(\mathcal{X}^{tr}))
\end{equation}
where $\hat{\textbf{y}}_j$ corresponds to the j-th predicted scene label and $\Phi = \{\phi_1,\phi_2...\phi_t\}$ are the set of different feature representations obtained from the $\textbf{x}_i$. This trained model should be able to correctly classify a set of unlabelled test samples $\mathcal{X}^{te} = \{\textbf{x}_1,\textbf{x}_2...\textbf{x}_M\}$. It is to be noted that while the goal of each of our five models is to perform place categorization, it is the $\Phi$ which varies across them. We now describe the two base modules, and how our five models are derived and trained from them. The complete network architecture is given in Figure \ref{full_architecture}. 

\input{images/full_architecture.tex}

\subsection{Scene Recognition}
For obtaining the scene gist, we use the PlacesCNN model. The base architecture is that of Resnet-18~\cite{he2016deep} which has been pretrained on the ImageNet dataset~\cite{deng2009imagenet} and then finetuned on the Places365 dataset~\cite{zhou2017places}. For each of our 5 models, we choose seven classes out of the total 365 classes, which are integral to the recognition of indoor home/office environments - Bathroom, Bedroom, Corridor, Dining room, Living room, Kitchen and Office. We use the official training and validation split provided for our work. The training set consists of 5,000 labelled images for each scene class, while the test set contains 100 images for each scene.

\subsection{Object Detection}
Object detection is a domain that has benefited immensely from the developments in deep learning. Recent years have seen people develop many algorithms for object detection, some of which include YOLO~\cite{redmon2016you,redmon2017yolo9000,redmon2018:arxiv}, SSD~\cite{liu2016ssd}, Mask RCNN~\cite{he2017mask}, Cascade RCNN~\cite{cai2018cascade} and RetinaNet~\cite{lin2017focal}. In this paper, we work with the YOLOv3~\cite{redmon2018:arxiv} detector, mainly because of its speed, which makes real-time processing possible. It is a Fully Convolutional Network (FCN), and employs the Darknet-53 architecture which has 53 convolution layers, consisting of successive 3x3 and 1x1 convolutional layers with some shortcut connections. The network used in this paper has been pre-trained to detect the 80 object classes of the MS-COCO dataset~\cite{lin2014microsoft}.

\subsection{Place Categorization models}
\subsubsection{Scene Only} The first model which we use consists of only the pre-trained and fine-tuned PlacesCNN with a simple Linear Classifier on top of it. This model accounts for a holistic representation of a scene, without specifically being trained to detect objects. Thus, the feature vector for this model is given by $\Phi_{scene} = \phi_s$.
\subsubsection{Object Only} The second model acts a Scene classifier using only the information of detected objects. There is no separate training required here to identify the individual scene attributes. For this purpose, we create a codebook of the most common COCO-objects seen in all the seven scenes. This is shown in Table~\ref{tab:top_obj}. It is to be noted that every object has been associated to only one scene (for instance, \textit{"bed"} is only associated to the \textit{Bedroom}) for classification purpose, thereby making it a \textit{landmark}. Since most images of the \textit{Corridor} scene do not have a landmark object, the codebook links this class to the absence of any object. For this model, the feature representation is given by $\Phi_{obj} = \phi_{\{obj\}}$ where $\{obj\}$ is the set of objects detected in the image. 
\input{tables/table4.tex}
\subsubsection{Scene+Attention} In this model, we compute the activation maps for the given image of a scene, and using those, we try to visualize where the network has its focus during scene classification. It is to be noted that while this model also performs place categorization, in addition, it highlights the important regions (objects, object-object interactions etc.) of a scene from the classification point of view. From the output of the final block convolutional layer (layer 4) of the WideResnet architecture~\cite{zagoruyko2016wide}, we get 14x14 feature blobs which retain spatial information corresponding to the whole image. Our model is similar to the \textit{soft} attention mechanism of \cite{xu2015show} in that here too, we assign the weights to be the output of a softmax layer, thereby associating a probability distribution to it. However, since we are not classifying based on a sequence of images, we do not employ a recurrent network to compute the sequential features. Instead, we simply utilize the weights of the final FC layer and take its dot product with the feature blobs to obtain the heatmap. The final step is to upsample this 14x14 heatmap to the input image size, and then overlay it on top to obtain the activation mask $m(x_n)$ of the input image $x_n$. Therefore, the feature representation for this model is $\Phi_{attn.} = \phi_{m(x_n)}$. The basic architecture is given in Figure \ref{attention}. 
\input{images/attention.tex}
\subsubsection{Combined} In this model, we use the PlacesCNN mentioned above as a feature extractor to give the semantics of a scene. In addition, the YOLO detector gives us the information regarding the objects present in the image. Given an image of a scene, this model creates a one hot-encoded vector of 80 dimensions, corresponding to the object classes of MS-COCO, with only the indices of the detected objects set to 1. We then concatenate this vector along with that of the output of the scene feature extractor, and train a Linear Classifier on top of it. Since we combine the two different features of scene and objects, the feature representation here is given by $\Phi_{comb.} = \{\phi_s,\phi_{\{obj\}}\}$.
\subsubsection{Scene+N-best objects} Our final model is similar to the above in that here also, we use both the PlacesCNN and the YOLO detector. However, this model does not need to be retrained again and so, it is significantly faster. For this model, we place a certain confidence threshold on the scene detector, and only when the probability of classification is below this threshold, we search for the information about specific objects in the scene (as obtained from Table~\ref{tab:top_obj}). The reason for introducing this as a new model is two-fold. Firstly, we eliminate the scenario of looking at every object present since it is often redundant, given the semantics of the scene. Secondly, this is similar to how we as human beings operate when we come across an unknown scene. The feature representation for this model is given by $\Phi_{N-best} = \{\phi_s,\phi_{\{N-obj\}}\}$.
\vspace{-0.055in}

%% file: images/full_architecture.tex
\begin{figure}[htbp]
  \centering
  \includegraphics[width=0.47\textwidth]{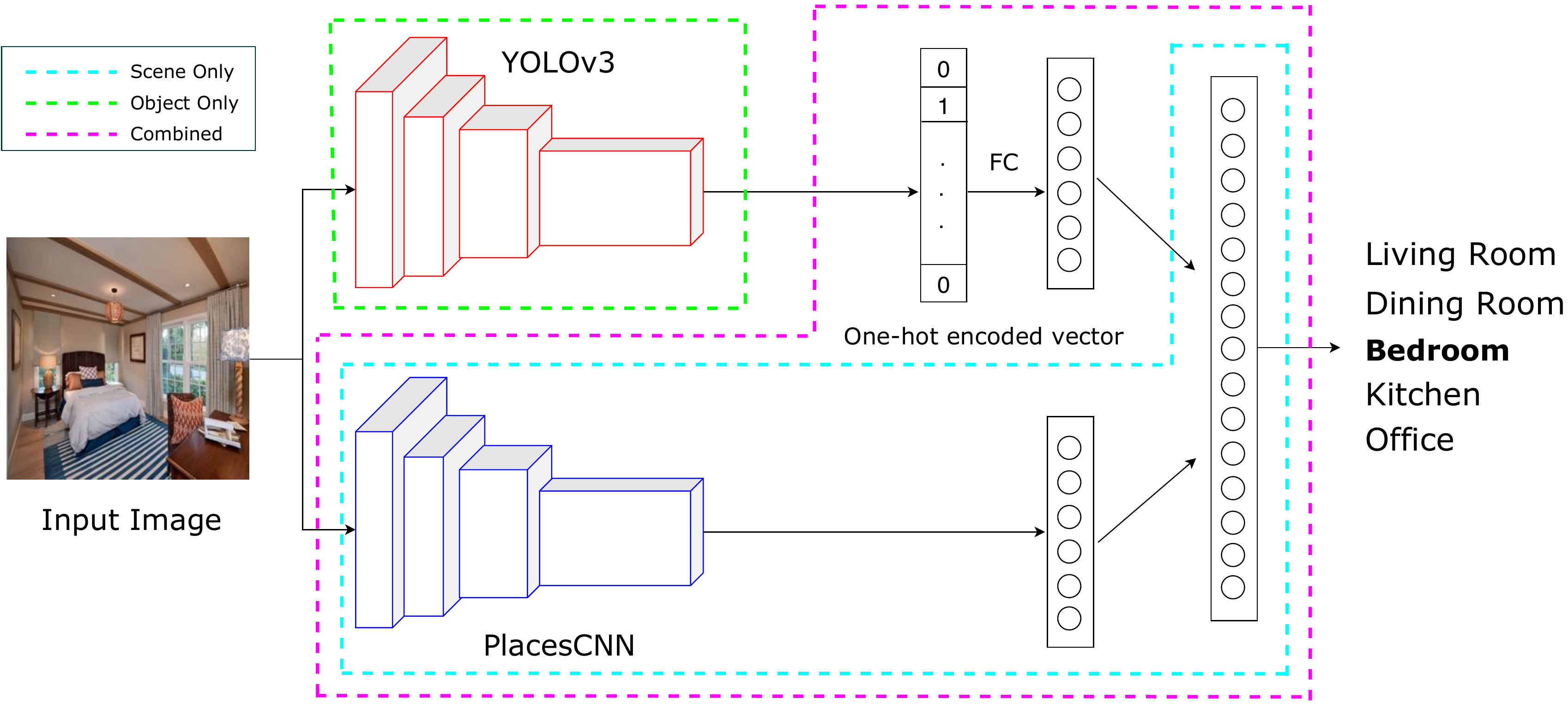}
  \caption{\small Model Architecture (best viewed in color). The highlighted regions represent the portion of the network which was trained for the respective models.}
  \label{full_architecture}
\end{figure}
\vspace{-0.2in}

%% file: tables/table4.tex
\begin{table}[htbp]
\caption{Top landmark objects (non-human) for the seven different scene classes}
\small
\centering
\scalebox{0.87}{
\begin{tabular}{|l||c|c|c|c|}
\hline
Bathroom & Toilet & Sink & - & -\\
\hline
Bedroom & Bed & - & - & -\\
\hline
Corridor & - & - & - & -\\
\hline
Dining Room & Dining Table & Wine Glass & Bowl & -\\
\hline
Kitchen & Oven & Microwave & Refrigerator & -\\
\hline
Living Room & Sofa & Vase & - & -\\
\hline
Office & TV-Monitor & Laptop & Keyboard & Mouse\\
\hline
\end{tabular}
}

\label{tab:top_obj}
\end{table}

%% file: images/attention.tex
\begin{figure}[htbp]
  \centering
  \includegraphics[width=\linewidth]{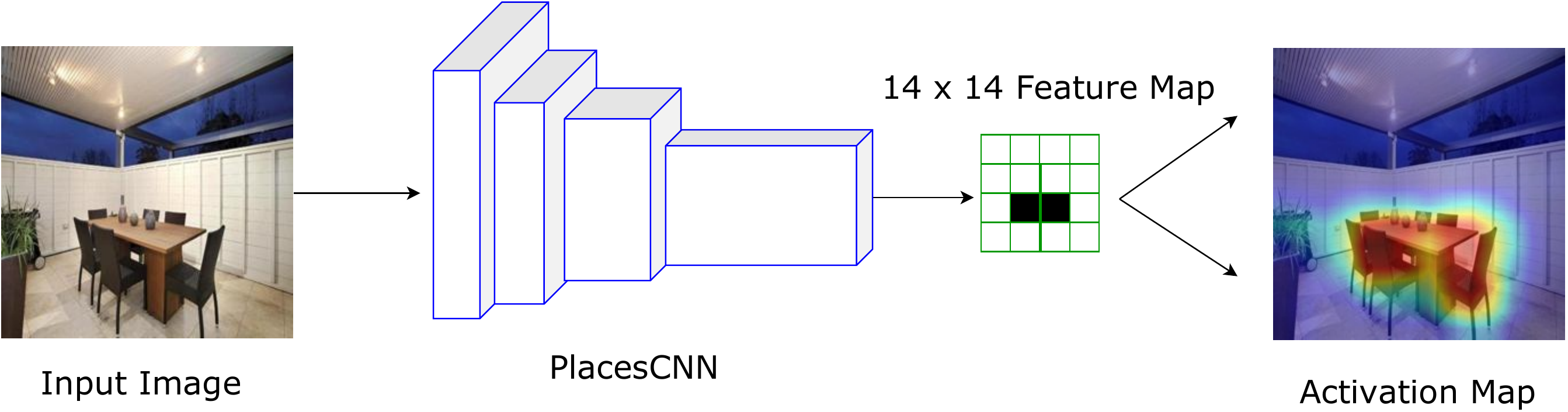}
  \caption{\small Architecture for Generation of the Activation Map (best viewed in color). The 14x14 feature maps obtained from the block layer 4 of WideResnet \cite{zagoruyko2016wide} are combined with the weights from the final FC layer, and then their dot product is upsampled to the image size and overlaid on top to get the activation maps}
  \label{attention}
\end{figure}

%% file: files/expresults.tex
\section{Experiments and Results}\label{sec:exp_res}

We evaluated our five models described above on a number of platforms. In this section, we first describe our training procedure, and then talk about the different experiment settings used for evaluation.
\subsection{Training Procedure}
As mentioned in Section \ref{sec:method}, the base architecture for our scene classifier is the ResNet-18 architecture. The data pre-processing and training process is similar to~\cite{zhou2017places}. We used the Stochastic Gradient Descent (SGD) optimizer with an initial learning rate of 0.1, momentum of 0.9, and a weight decay of 10$^{-4}$. For the $\Phi_{scene}$ and the $\Phi_{attn.}$ models, the training was performed for 90 epochs with the learning rate being decreased by a factor of 10 every 30 epochs. The $\Phi_{comb.}$ model converged much faster and so, it was only trained for 9 epochs, with the learning rate reduced by 10 times after every 3 epochs. For all the 3 training procedures, the cross-entropy loss function was optimized, which minimizes the cost function given by
\begin{equation}
    J(\hat{\textbf{y}},\textbf{y}) = -\frac{1}{N}(\sum_{j=1}^{N}\textbf{y}_j\odot log(\hat{\textbf{y}}_j))
\end{equation}

The training process was carried out on an NVIDIA Titan Xp GPU using the PyTorch framework. The performance of the five DEDUCE algorithms on the test set of Places365 is shown in Table~\ref{tab:places} for the seven classes chosen.
\input{tables/table0.tex}
\subsection{Experiment Settings}


In order to check the robustness of our models, we further evaluated their performance on two state-of-the-art still-image datasets.

\subsubsection{SUN Dataset}
The SUN-RGBD dataset~\cite{song2015sun}  is one of the most challenging scene understanding datasets in existence. It consists of 3,784 images using Kinect v2 and 1,159 images using Intel RealSense cameras. In addition, there are 1,449 images from the NYUDepth V2~\cite{silberman2012indoor}, and 554 manually selected realistic scene images from the Berkeley B3DO Dataset~\cite{janoch2013category}, both captured  by  Kinect  v1. Finally, it has 3,389 manually selected distinguished frames without significant motion blur from the SUN3D videos~\cite{xiao2013sun3d} captured by Asus Xtion. Out of this, we sample the seven classes of importance and use the official test split to evaluate our models. We only consider the RGB images for this work since our training data doesn't have depth information. The performance is summarized in Table \ref{tab:sun}. 
\input{tables/table1.tex}

Upon comparison with Table \ref{tab:places}, which contains the results on the Places365 dataset where our models were fine-tuned, a number of observations can be made which are consistent for both the datasets. Firstly, the $\Phi_{comb.}$ model performs the best. This is intuitive since here, the scene classification is done using the combined training of both the information about the scene attributes and the object identity. Secondly, the $\Phi_{obj}$ model works the best for the \textit{Dining Room} class, even though its overall performance is the worst. This trend can be attributed to the fact that dining rooms can be easily identified by the presence of specific objects, whereas the scene attributes might throw in some confusion (for instance when the kitchen/living room is partially visible in the image of a dining room). Thirdly, for \textit{Corridor}, the performance of the $\Phi_{attn.}$ model is best for both the datasets. This supports the fact that in order to classify a scene like a corridor, viewing only a small portion of the image close to the vanishing point is sufficient. Finally, the $\Phi_{N-best}$ model performs just as good or better than the $\Phi_{scene}$ model. This proves that presence of objects does indeed improve the scene classification. For the best performance using the $\Phi_{N-best}$ model, the threshold was set to 0.5 for the \textit{Places} dataset while it was 0.6 for the \textit{SUN} dataset. The reason for the higher confidence on scene attributes for \textit{Places} dataset is most likely due to the fact that the scene classifier itself was fine tuned on it.

\subsubsection{VPC Dataset}
The Visual Place Categorization dataset~\cite{Wu2009:IROS} consists of videos captured autonomously using a HD camcorder (JVC GR-HD1) mounted on a rolling tripod. The data has been collected from 6 different home environments, and three different floor types. The advantage of this dataset is that the collected data closely mimics that of the motion of a robot - instead of focusing on captured frames or objects/furniture in the rooms, the operator recording the data just traversed across all the areas in a room while avoiding collision with obstacles. For comparison with the state-of-the-art algorithms, we test our methods only on the five classes which are present in all the homes - Bathroom, Bedroom, Dining room, Living room and Kitchen. Table~\ref{tab:sota_class} contains the results for the individual home environments for these five classes. For the AlexNet~\cite{krizhevsky2012imagenet} and ResNet~\cite{he2016deep} models, we adopt the same training procedure as in \cite{mancini2018robust}. It can be seen from the table that our models perform better than the rest in all but one of the home environments and much better in the overall performance. 
\input{images/realworldvid.tex}

\input{tables/table2.tex}
\input{tables/table3.tex}

Table \ref{tab:sota} further compares our models with all other baseline algorithms tested on the VPC dataset. The reported accuracies are the average over all the six home environments. We first consider the methods described in~\cite{Wu2009:IROS}, which use SIFT and CENTRIST features with a Nearest Neighbor Classifier, and also exploit temporal information between images by coupling them with Bayesian Filtering (BF). Next, we look at the approach of~\cite{fazl2012histogram} where Histogram of Oriented Uniform Patterns (HOUP) is used as input to the same classifier. \cite{yang2012object} proposed the method of using object templates for visual place categorization, and reported results for Global configurations approach with Bayesian Filtering (G+BF), and that combined with the object templates (G+O(SIFT)+BF). Ushering the deep learning era, AlexNet~\cite{krizhevsky2012imagenet} and ResNet~\cite{he2016deep} architectures give better results, both with their base models, as well as the Batch Normalized (BN) and the Weighted Batch Normalized versions~\cite{mancini2018robust}. However, comparisons with our $\Phi_{scene}$ and $\Phi_{comb.}$ models show that our methods beat all the other results by significant margins.

\subsubsection{Real-World Scene Recognition}
Since most real-world data for robotics research occur in the form of a sequence of images, we also perform rigorous experiments in domains beyond the aforementioned still image datasets. Figure~\ref{real_world_vids} shows the results of scene recognition on real-world data recorded using hand-held cameras. We employ the $\Phi_{N-best}$ model for these cases due to its ability to mimic the natural behavior of humans, whereby an initial prediction is made based on the scene attributes, and if unsure, more information related to specific objects is considered in order to update/re-inforce the initial prediction. The top row corresponds to the tour of a semi-furnished real estate home obtained from YouTube which only has the relevant objects in the scene. Although it is not a continuous tour of the house, it does contain all the rooms in a sequential data format. Also, professional photographers captured this video and hence, the image quality and white-balance of the camera is pretty good. The next-2 rows pose a more challenging case as they correspond to homes currently inhabited by people. We consider two examples of these houses, one which is a standard bungalow residence, while the other being a student apartment. From experience, the bungalow is a much cleaner home, whereas student apartments are prone to presence of cluttered objects and overlapping scene boundaries. Moreover, the videos were recorded by the inhabitants using their cellphone cameras. This inherently brings motion blur into the picture, especially during scene transitions. Finally, the last row depicts the settings of a house from the movie \textit{"Father of the Bride"}. This ensures that our model is robust enough to classify scenes even when the focus of the recording is on people instead of the background settings. 

\subsubsection{Semantic Mapping}
The experimental setting for semantic mapping involves running our algorithm on a mobile robot platform in two different environments of the UC San Diego campus. The platform is a Fetch Mobile Manipulator and Freight Mobile Robot Base by Fetch Robotics\footnote{\url{https://fetchrobotics.com/robotics-platforms/}}. Figure~\ref{fig:fetch_semMapping} shows the robot performing scene classification in one of the environments. 

The semantic maps for the experiments were constructed using OmniMapper~\cite{Trevor2014:ICRA}. It utilizes the GTsam library in order to optimize a graph of measurements between robot poses along a trajectory, and between robot poses and various landmarks in the environment. The measurements of simple objects like points, lines, and planes are data associated with mapped landmarks using the joint compatibility branch and bound (JCBB) technique~\cite{neira2001data}. The regions for color segmentation are acquired by the Gaussian Region algorithm of~\cite{nieto2010:IROS}. However, in~\cite{nieto2010:IROS}, the map partitions were built through human guidance, whereby the robot was taken on a tour of the space (either by driving the robot manually, or using a person following behavior) and the respective scene labels were taught to it. This is in contrast to our approach, where the labels are learned from our visual place categorization system. Thus, the robot is itself capable of identifying the scenes without any human guide. We used the $\Phi_{N-best}$ model for this task, and retrained the scene classifier to exclude the \textit{Bedroom}, \textit{Dining Room} \& \textit{Bathroom} scenes, and instead include \textit{Conference Room} as it is more likely to occur in an academic building environment. 

Figure \ref{map:cse} shows navigation of the robot in the Computer Science and Engineering (CSE) building. Our system was able to classify the five regions of the floor map. However, there are some regions detected by OmniMapper using the laser range finder. These are painted in white to denote their invisibility to the camera. The second test environment is the Contexual Robotics Institute (CRI) building, which has a very different floor map in comparison to CSE. The result of the run made here is shown in Figure \ref{maps:atkinson}.

%% file: tables/table0.tex
\begin{table}[htbp]
\caption{Accuracy in percentage of DEDUCE on Places365 dataset}
\small
\centering
\scalebox{0.9}
{
\begin{tabular}{|l||c|c|c|c|c|}
\hline
Scenes & $\Phi_{scene}$ & $\Phi_{obj}$ & $\Phi_{attn.}$ & $\Phi_{comb.}$ & $\Phi_{N-best}$\\ 
\hline\hline
Dining room & 79 & \textbf{94} & 75 & 79 & 80 \\
Bedroom & 90 & 74 & 90 & 90 & \textbf{91} \\
Bathroom & \textbf{92} & 65 & \textbf{92} & 91 & \textbf{92} \\
Corridor & 94 & 90 & \textbf{99} & 96 & 94 \\
Living Room & \textbf{84} & 25 & 68 & 80 & \textbf{84} \\
Office & 85 & 29 & 76 & \textbf{94} & 83 \\
Kitchen & \textbf{87} & 62 & 70 & \textbf{87} & \textbf{87} \\
\hline
\textbf{Avg} & 87.3 & 62.6 & 81.4 & \textbf{88.1} & 87.3\\
\hline
\end{tabular}
}
\label{tab:places}
\end{table}

%% file: tables/table1.tex
\begin{table}[h]
\renewcommand\thetable{III}
\caption{Accuracy in percentage of DEDUCE on SUN dataset}
\small
\centering
\scalebox{0.9}
{
\begin{tabular}{|l||c|c|c|c|c|}
\hline
Scenes & $\Phi_{scene}$ & $\Phi_{obj}$ & $\Phi_{attn.}$ & $\Phi_{comb.}$ & $\Phi_{N-best}$\\ \hline\hline
Dining room  & 65.2 & \textbf{83.7} & 53.3 & 67.4 & 72.8\\
Bedroom & 43.7 & 36.5 & \textbf{48.9} & \textbf{48.9} & 47.3\\
Bathroom  & 94.5 & 87.0 & \textbf{97.3} & 96.6 & 95.2\\
Corridor & 44.4 & \textbf{67.6} & \textbf{67.6} & 44.4 & 41.7\\
Living Room & 58.8 & 24.0 & 43.6 & \textbf{59.2} & 58.8\\
Office & 84.0 & 12.6 & 75.8 & \textbf{90.6} & 80.6\\
Kitchen & 77.1 & 63.5 & 63.9 & \textbf{83.8} & 77.4\\
\hline
\textbf{Avg} & 66.8 & 53.6 & 64.3 & \textbf{70.1} & 67.7\\
\hline
\end{tabular}
}
\label{tab:sun}
\end{table}

%% file: images/realworldvid.tex
\begin{figure*}[!b]
  \centering
  \includegraphics[width=0.85\textwidth]{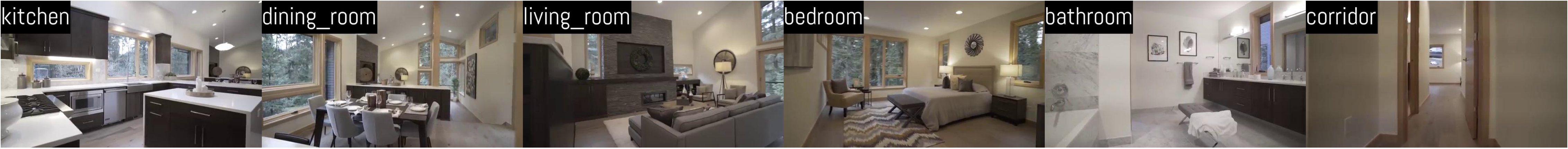}
  \includegraphics[width=0.85\textwidth]{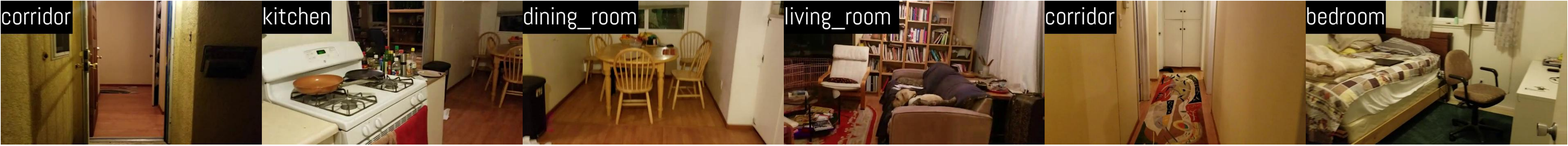}
  \includegraphics[width=0.85\textwidth]{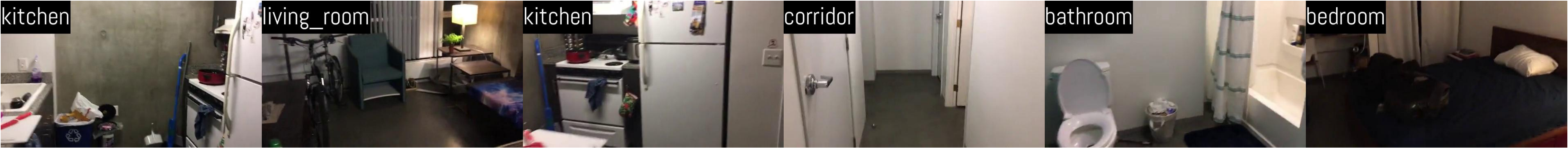}
  \includegraphics[width=0.85\textwidth]{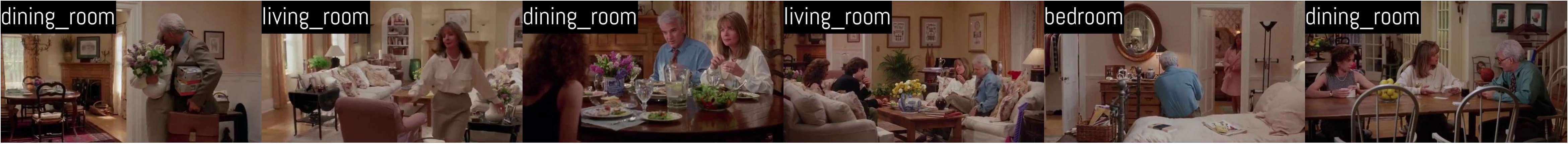}
  \caption{\small Detection Results on Real-World Videos. Top row corresponds to the video of a Real-Estate model house. The next 2 rows are from the houses of the authors and their friends. The bottom row is obtained from a house in the movie \textit{"Father of the Bride"}.}
  \label{real_world_vids}
\end{figure*}

%% file: tables/table2.tex
\begin{table}[ht]
\renewcommand\thetable{IV}
\caption{VPC Dataset: Average Accuracy across the 6 home environments}
\small
\centering
\scalebox{0.85}{
\begin{tabular}{|l||c|c|c|c|c|c|c|}
\hline
 Networks & H1 & H2 & H3 & H4 & H5 & H6 & avg.\\

\hline\hline
AlexNet & 49.8 & 53.4 & 49.2 & 64.4 & 41.0 & 43.4 & 50.2\\
AlexNet+BN & 54.5 & 54.6 & 55.6 & 69.7 & 41.8 & 45.9 & 53.7\\
AlexNet+WBN & 54.7 & 51.9 & 61.8 & 70.6 & 43.9 & 46.5 & 54.9\\
AlexNet+WBN$^*$ & 53.5 & 54.6 & 55.7 & 68.1 & 44.3 & 49.9 & 54.3\\
ResNet & 55.8 & 47.4 & 64.0 & 69.9 & 42.8 & 50.4 & 55.0\\
ResNet+WBN & 55.7 & 49.5 & \textbf{64.7} & 70.2 & 42.1 & 52.0 & 55.7\\
ResNet+WBN$^*$ & 56.8 & 50.9 & 64.1 & 69.3 & 45.1 & 51.6 & 56.5\\
\textbf{Ours ($\Phi_{scene}$)} & \textbf{63.7} & 57.3 & 63.7 & \textbf{71.4} & 60.2 & 65.9 & 63.7\\
\textbf{Ours ($\Phi_{comb.}$)} & \textbf{63.7} & \textbf{60.7} & 64.5 & 70.7 & \textbf{65.7} & \textbf{68.8} & \textbf{65.7}\\
\hline
\end{tabular}
}
\label{tab:sota_class}
\end{table}

%% file: tables/table3.tex

\begin{table*}[!t]
\renewcommand\thetable{V}
\caption{VPC Dataset: Comparison with state-of-the-art}
\scriptsize
\centering
\scalebox{0.97}{
\begin{tabular}{|l||c|c|c|c|c|c|c|c|c|c|c|c|c|c|}
\hline
Method & \multicolumn{4}{c|}{\cite{Wu2009:IROS}} & \cite{fazl2012histogram} & \multicolumn{2}{c|}{\cite{yang2012object}} & \multicolumn{3}{c|}{\text{AlexNet}}  &\multicolumn{2}{c|}{\text{ResNet}} &  \multicolumn{2}{c|}{\textbf{Ours}} \\
\hline
Config. & SIFT & SIFT+BF & CE & CE+BF & HOUP & G+BF & G+O(SIFT)+BF & Base & BN & WBN$^*$ & BN & WBN$^*$ & \textbf{Scene-only} & \textbf{Combined}\\
\hline\hline
Acc. & 35.0 & 38.6 & 41.9 &  45.6 & 45.9 & 47.9 & 50 & 50.2 & 53.7 & 54.3 & 55.0 & 56.5 & \textbf{63.7} & \textbf{65.7} \\
\hline
\end{tabular}
}
\label{tab:sota}
\end{table*}

%% file: files/conclusion.tex
\section{Conclusion \& Future Work} \label{sec:conclusion}
In this paper, we considered five different models for place categorization, which are derived mainly from two base modules - a scene recognizer, and an object detector. We demonstrated the effectiveness of our algorithms in a series of experiments, ranging from scene recognition in still-image data sets to real-world videos captured from different sources, and finally via the generation of labeled semantic maps using data gathered by multiple mobile robot platforms. 

We showed that (i) different models are favorable for different scenes (Table \ref{tab:places} and \ref{tab:sun}), and thus the ideal scene recognition system would likely be a combination of these five models, (ii) the proposed methods give successful results on many different types of video recordings, even when they are prone to object clutter, motion blur, and overlapping boundaries and (iii) our models are robust enough to be tested on data gathered by mobile robotic platforms on multiple building scenarios which are affected by occlusions and poor lighting conditions.

We plan to extend our experimental evaluation to other mobile robots and then flying robots. While we demonstrated the effectiveness of our approach in different indoor scenarios, we believe that the next step is to allow the robot to walk on a tour and label important regions in its environment. This would be useful for navigation in unknown places using a semantic map for robots and humans. Furthermore, our system could be applied to autonomous robots, thus enabling them to assist humans in safety and rescue missions inside a house or a building.
\input{images/semmaps.tex}
\input{tables/table5.tex}
\input{tables/table6.tex}
\vspace{-0.07in}

%% file: images/semmaps.tex
\begin{figure}[!ht]
\centering
    \begin{subfigure}[b]{0.6\columnwidth}
    \centering
    \includegraphics[width=.8\textwidth]{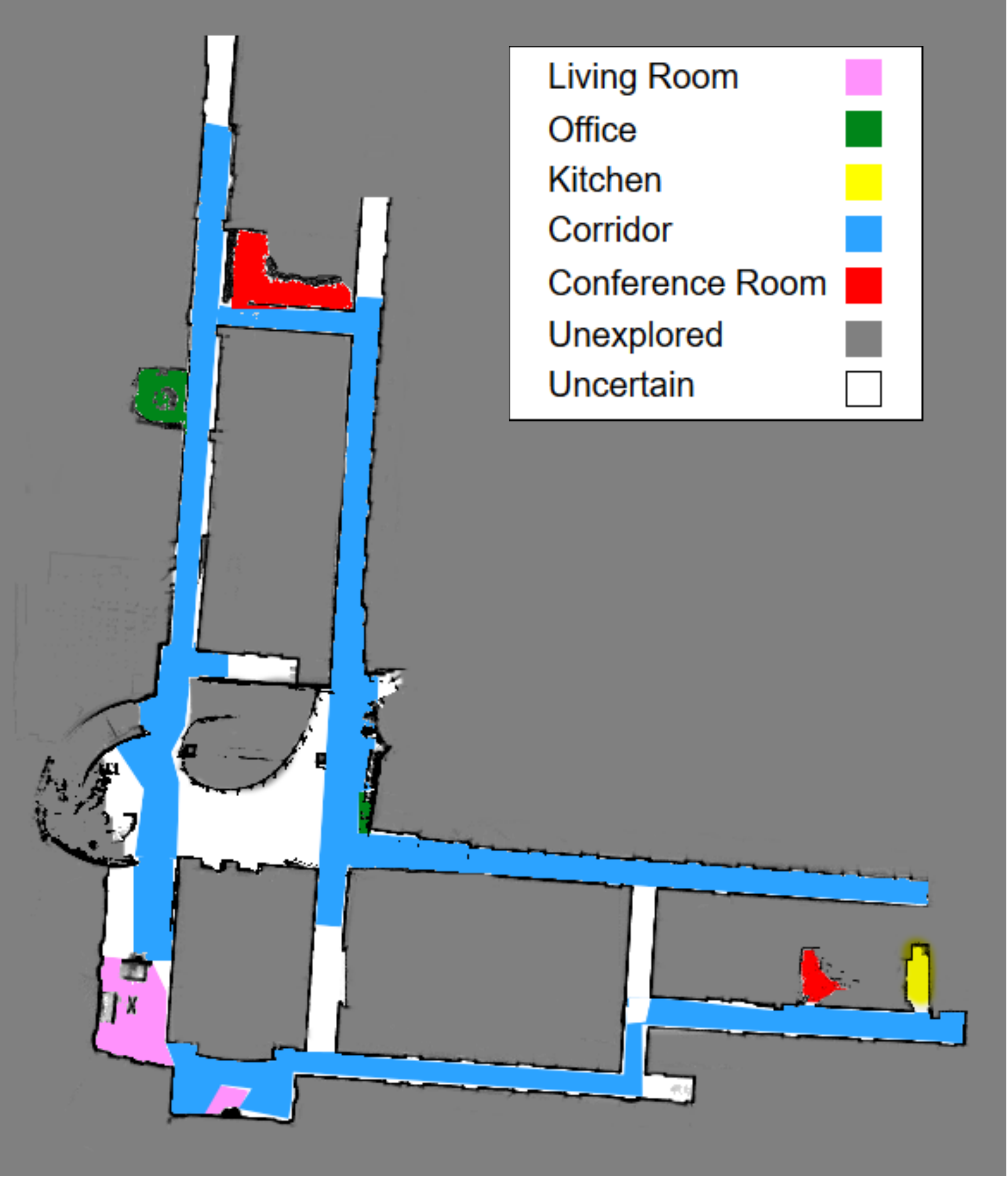}
    \caption{\small Semantic Map of the CSE building}
    \label{map:cse}
    \end{subfigure}

    \begin{subfigure}[b]{0.6\columnwidth}
    \centering
  \includegraphics[width=.8\textwidth]{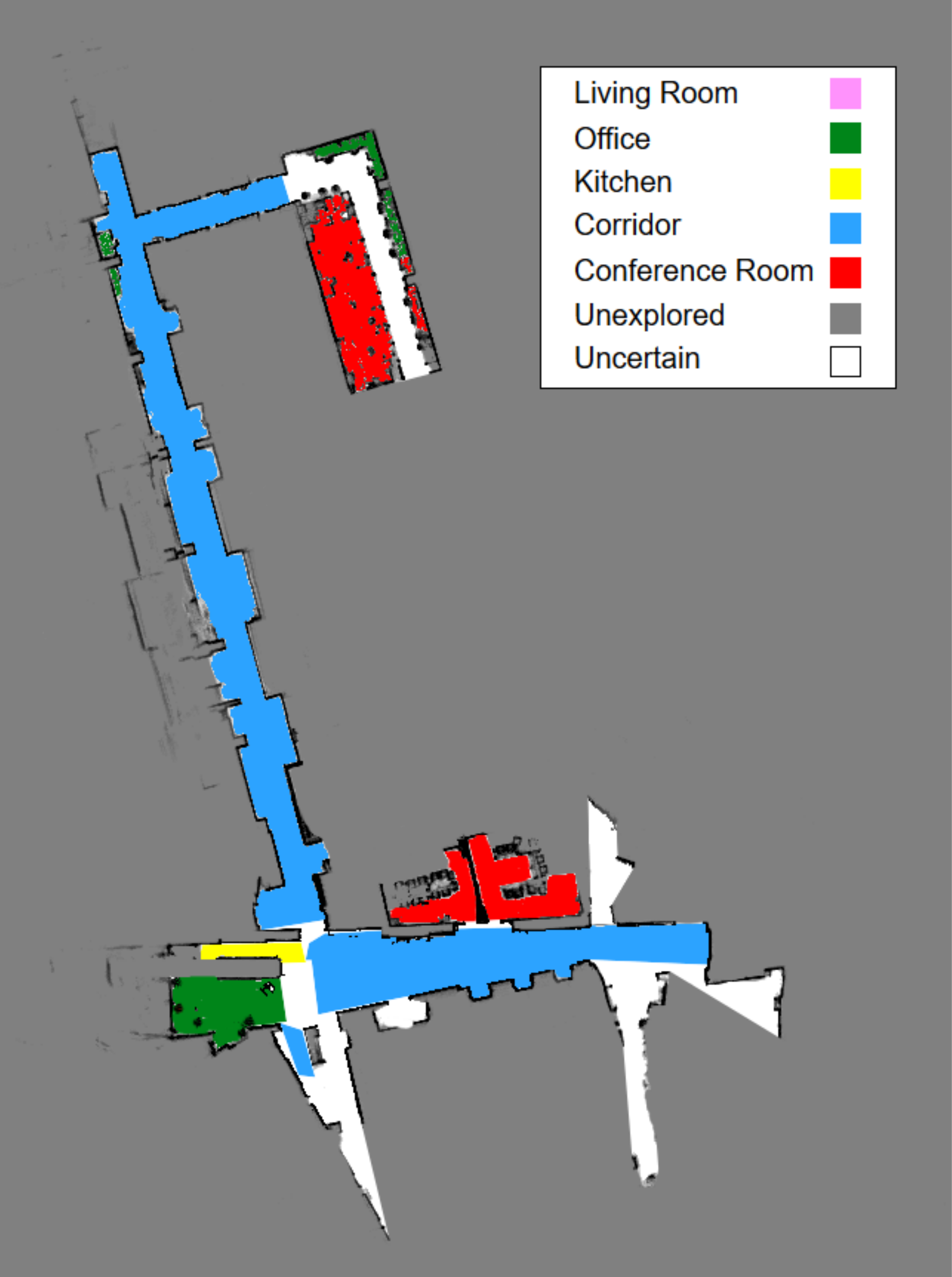}
  \caption{\small Semantic Map of the CRI}
  \label{maps:atkinson}
    \end{subfigure}

\caption{\small Place categorization experiments with mobile robots (best viewed in color). Each color represents one of the seven classes of the visual place categorization that our system classified.}
\label{fig:semmaps}
\end{figure}

%% file: tables/table5.tex
\begin{table*}[t]
\caption{VPC Dataset: $\Phi_{scene}$ model results for individual homes}
\scriptsize
\centering
\scalebox{0.95}{
    \begin{tabular}{|l||c|c|c|c|c|c|c|c|c|c|c|c|c|c|c|}
    \hline
     & \multicolumn{3}{c|}{Home 1} & \multicolumn{3}{c|}{Home 2} & Home 3 & \multicolumn{2}{c|}{Home 4} & \multicolumn{2}{c|}{Home 5} & \multicolumn{2}{c|}{Home 6} & \textbf{Avg. (ours)} & Avg. \cite{fazl2012histogram}\\
    \hline
    House number & 0 & 1 & 2 & 0 & 1 & 2 & 1 & 1 & 2 & 1 & 2 & 0 & 1 & &\\
    \hline\hline
    Dining Room & - & 71.5 & - & - & 43.5 & - & 72.4 & 49.4 & - & 19.6 & - & - & 52.0 & \textbf{51.4} & 15.5\\
    Bedroom & - & 32.0 & 43.1 & 28.8 & - & 41.1 & 41.7 & 60.7 & 45.7 & - & 51.5 & 43.5 & 50.3 & 43.8 & \textbf{73.0}\\
    Bathroom & 83.4 & 63.6 & 50.2 & 85.0 & - & 86.5 & 79.0 & 96.1 & 91.6 & 96.3 & 59.6 & 84.3 & 67.0 & \textbf{78.6} & 68.7\\
    Living Room & - & 97.6 & - & - & 37.4 & - & 72.9 & 79.0 & - & 76.2 & - & - & 77.6 & \textbf{73.5} & 25.9\\
    Kitchen & - & 40.0 & - & - & 72.5 & - & 52.3 & 85.7 & - & 67.5 & - & - & 92.8 & \textbf{68.5} & 46.6\\
    \textbf{Avg.} & \textbf{83.4} & \textbf{61.0} & \textbf{46.7} & \textbf{56.9} & \textbf{51.1} & \textbf{63.8} & \textbf{63.7} & \textbf{74.2} & \textbf{68.6} & \textbf{64.9} & \textbf{55.5} & \textbf{63.9} & \textbf{68.0} & &\\
    \hline
    \end{tabular}
}
\label{tab:scene_vpc_all_classes}
\end{table*}

%% file: tables/table6.tex
\begin{table*}[!b]
\caption{VPC Dataset: $\Phi_{comb.}$ model results for individual homes}
\scriptsize
\centering
\scalebox{0.95}{
\begin{tabular}{|l||c|c|c|c|c|c|c|c|c|c|c|c|c|c|c|c|}
\hline
 & \multicolumn{3}{c|}{Home 1} & \multicolumn{3}{c|}{Home 2} & Home 3 & \multicolumn{2}{c|}{Home 4} & \multicolumn{2}{c|}{Home 5} & \multicolumn{2}{c|}{Home 6} & \textbf{Avg.(ours)} & Avg. \cite{fazl2012histogram}\\
\hline
House number & 0 & 1 & 2 & 0 & 1 & 2 & 1 & 1 & 2 & 1 & 2 & 0 & 1 & &\\
\hline\hline
Dining Room & - & 83.9 & - & - & 47.7 & - & 79.8 & 48.8 & - & 25.00 & - & - & 47.1 & \textbf{55.4} & 15.5\\
Bedroom & - & 32.2 & 37.4 & 31.9 & - & 49.6 & 53.0 & 66.2 & 41.7 & - & 56.6 & 51.8 & 58.8 & 47.9 & \textbf{73.0}\\
Bathroom & 82.4 & 58.9 & 48.3 & 86.4 & - & 82.1 & 79.9 & 91.8 & 90.5 & 96.6 & 63.3 & 83.0 & 64.9 & \textbf{77.3} & 68.7\\
Living Room & - & 97.6 & - & - & 49.7 & - & 54.2 & 82.0 & - & 81.5 & - & - & 88.1 & \textbf{75.5} & 25.9\\
Kitchen & - & 56.0 & - & - & 73.7 & - & 55.6 & 87.9 & - & 83.2 & - & - & 92.3 & \textbf{74.8} & 46.6\\
Avg. & \textbf{82.4} & \textbf{65.7} & \textbf{42.9} & \textbf{59.1} & \textbf{57.0} & \textbf{65.9} & \textbf{64.5} & \textbf{75.4} & \textbf{66.1} & \textbf{71.6} & \textbf{59.9} & \textbf{67.4} & \textbf{70.3} & &\\
\hline
\end{tabular}
}
\label{tab:combined_vpc_all_classes}
\end{table*}

%% file: files/acknowledgements.tex
\section*{Acknowledgements} \label{sec:acknowledgements}

The authors would like to thank Army Research Laboratory (ARL) W911NF-10-2-0016 Distributed and Collaborative Intelligent Systems and Technology (DCIST) Collaborative Technology Alliance for supporting this research.

%% file: files/appendix.tex
 \section*{Appendix}
Here, we show the detailed results on the individual floor environments of each home of the VPC dataset \cite{Wu2009:IROS}. 

Table \ref{tab:scene_vpc_all_classes} shows the results for the $\Phi_{scene}$ model while Table \ref{tab:combined_vpc_all_classes} for the $\Phi_{comb.}$ model. Since all the scenes are not present in some of the floors, some of the cells are kept empty. We compare the average accuracy obtained per scene over the 6 houses, and compare the result with that reported in \cite{fazl2012histogram}. Clearly, both our algorithms outperform the reported score by a long margin, with the only exception of the bathroom class. Comparing our two models, we still see that the $\Phi_{comb.}$ model performs better than the $\Phi_{scene}$ model. 
In Figure \ref{acc}, we show the rate of convergence of the $\Phi_{scene}$ and the $\Phi_{comb.}$ models. It can be observed that the latter model converged much faster as compared to the former. This proves the worth of adding object information to scene classifier. However, it should be noted that the training process is much slower for $\Phi_{comb.}$. This is expected since here, the outputs of two CNNs are concatenated and then retrained together. The convergence rate of the $\Phi_{attn.}$ model is similar to the $\Phi_{scene}$ model, and so we omitted it from Figure \ref{acc} to avoid clutter.
\vspace{-0.18in}
\input{images/acc.tex}

%% file: images/acc.tex
\begin{figure}[!h]
  \centering
  \includegraphics[width=0.75\linewidth]{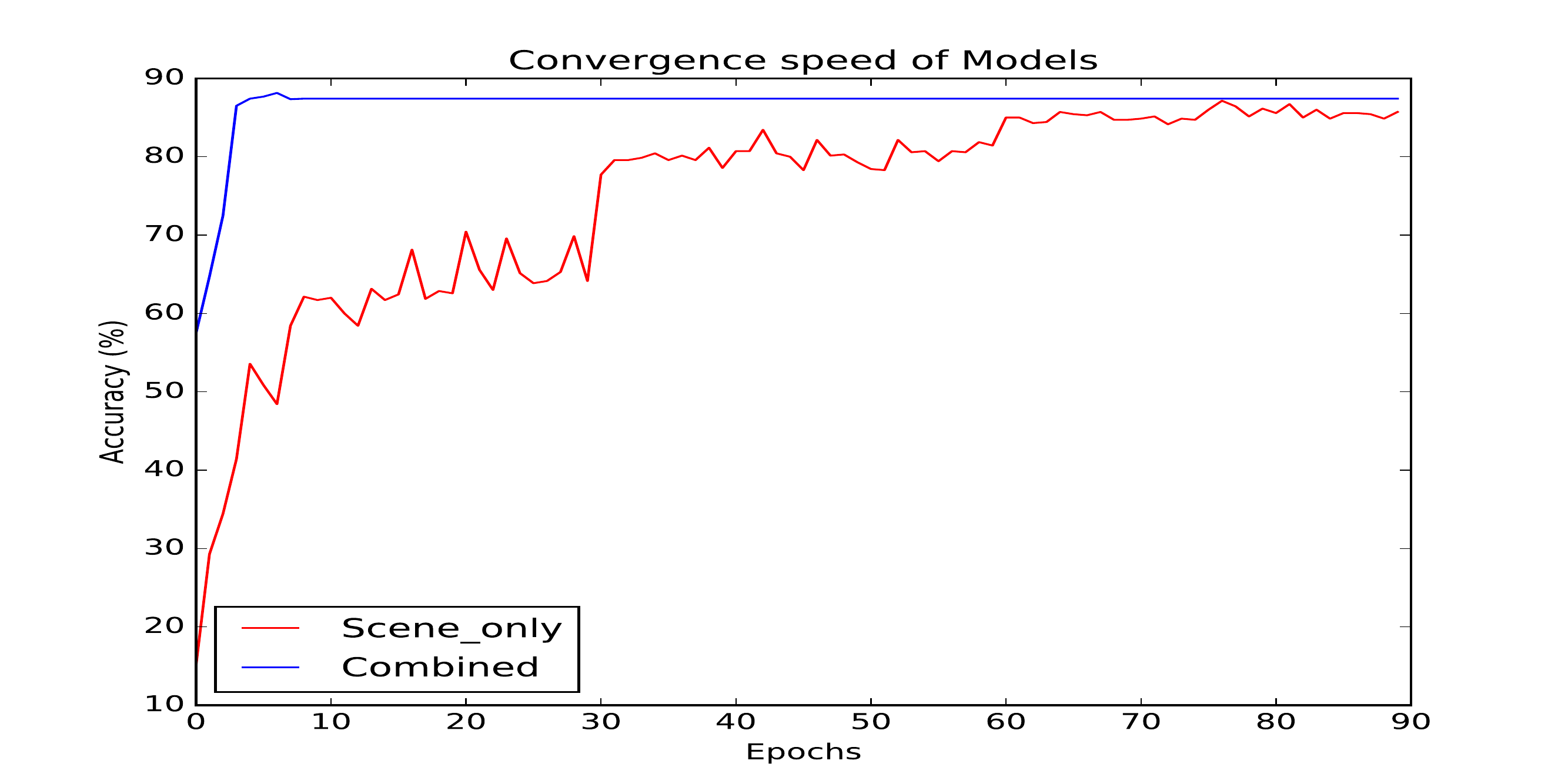}
  \caption{\small Convergence speed of models}
  \label{acc}
\end{figure}

%% file: iros2019_PlaceCat.bbl
\begin{thebibliography}{10}
\providecommand{\url}[1]{#1}
\csname url@samestyle\endcsname
\providecommand{\newblock}{\relax}
\providecommand{\bibinfo}[2]{#2}
\providecommand{\BIBentrySTDinterwordspacing}{\spaceskip=0pt\relax}
\providecommand{\BIBentryALTinterwordstretchfactor}{4}
\providecommand{\BIBentryALTinterwordspacing}{\spaceskip=\fontdimen2\font plus
\BIBentryALTinterwordstretchfactor\fontdimen3\font minus
  \fontdimen4\font\relax}
\providecommand{\BIBforeignlanguage}[2]{{%
\expandafter\ifx\csname l@#1\endcsname\relax
\typeout{** WARNING: IEEEtran.bst: No hyphenation pattern has been}%
\typeout{** loaded for the language `#1'. Using the pattern for}%
\typeout{** the default language instead.}%
\else
\language=\csname l@#1\endcsname
\fi
#2}}
\providecommand{\BIBdecl}{\relax}
\BIBdecl

\bibitem{Kostavelis2015:RAS}
I.~Kostavelis and A.~Gasteratos, ``Semantic {M}apping for {M}obile {R}obotics
  {T}asks,'' \emph{Robot. Auton. Syst.}, vol.~66, no.~C, Apr. 2015.

\bibitem{Bormann2016:ICRA}
R.~Bormann, F.~Jordan, W.~Li, J.~Hampp, and M.~H\"{a}gele, ``Room segmentation:
  Survey, implementation, and analysis.'' in \emph{ICRA}, 2016.

\bibitem{Niko2018:IJRR}
N.~S\"{a}nderhauf, O.~Brock, W.~Scheirer, R.~Hadsell, D.~Fox, J.~Leitner,
  B.~Upcroft, P.~Abbeel, W.~Burgard, M.~Milford, and P.~Corke, ``The limits and
  potentials of deep learning for robotics,'' \emph{IJRR}, 2018.

\bibitem{Torralba2003:ICCV}
A.~Torralba, K.~P. Murphy, W.~T. Freeman, and M.~A. Rubin, ``Context-based
  vision system for place and object recognition,'' \emph{ICCV}, vol.~1, p.
  273, 2003.

\bibitem{Wu2009:IROS}
J.~Wu, H.~I. Christensen, and J.~M. Rehg, ``Visual place categorization:
  Problem, dataset, and algorithm,'' in \emph{IROS}, 2009.

\bibitem{Pronobis2008:ICRA}
A.~Pronobis, O.~M. Mozos, and B.~Caputo, ``Svm-based discriminative
  accumulation scheme for place recognition,'' in \emph{ICRA}, 2008.

\bibitem{Siagian2007:PAMI}
C.~Siagian and L.~Itti, ``Rapid biologically-inspired scene classification
  using features shared with visual attention,'' \emph{T. PAMI}, vol.~29,
  no.~2, pp. 300--312, 2007.

\bibitem{oliva2001:IJCV}
A.~Oliva and A.~Torralba, ``Modeling the shape of the scene: A holistic
  representation of the spatial envelope,'' \emph{International Journal of
  Computer Vision}, vol.~42, pp. 145--175, 2001.

\bibitem{Quattoni2009:CVPR}
A.~Quattoni and A.~B. Torralba, ``Recognizing indoor scenes,'' in \emph{CVPR},
  2009, pp. 413--420.

\bibitem{Ekvall2007:Robotica}
S.~Ekvall, D.~Kragic, and P.~Jensfelt, ``Object detection and mapping for
  service robot tasks,'' \emph{Robotica}, vol.~25, no.~2, pp. 175--187, 2007.

\bibitem{tong2017sceneslam}
Z.~Tong, D.~Shi, and S.~Yang, ``Sceneslam: A {SLAM} framework combined with
  scene detection,'' in \emph{ROBIO}, 2017.

\bibitem{Espinace2010:ICRA}
P.~Espinace, T.~Kollar, A.~Soto, and N.~Roy, ``Indoor scene recognition through
  object detection,'' in \emph{ICRA}, 2010.

\bibitem{Kollar2009:ICRA}
T.~Kollar and N.~Roy, ``Utilizing object-object and object-scene context when
  planning to find things,'' in \emph{ICRA}, 2009, pp. 4116--4121.

\bibitem{liao2016understand}
Y.~Liao, S.~Kodagoda, Y.~Wang, L.~Shi, and Y.~Liu, ``Understand scene
  categories by objects: A semantic regularized scene classifier using
  convolutional neural networks,'' in \emph{ICRA}.\hskip 1em plus 0.5em minus
  0.4em\relax IEEE, 2016.

\bibitem{liao2017place}
------, ``Place classification with a graph regularized deep neural network,''
  \emph{IEEE Transactions on Cognitive and Developmental Systems}, vol.~9,
  no.~4, 2017.

\bibitem{sun2018scene}
H.~Sun, Z.~Meng, P.~Y. Tao, and M.~H. Ang, ``Scene recognition and object
  detection in a unified convolutional neural network on a mobile
  manipulator,'' in \emph{ICRA}.\hskip 1em plus 0.5em minus 0.4em\relax IEEE,
  2018, pp. 1--5.

\bibitem{Luo2018HierarchicalSM}
R.~C. Luo and M.~Chiou, ``Hierarchical semantic mapping using convolutional
  neural networks for intelligent service robotics,'' \emph{IEEE Access},
  vol.~6, pp. 61\,287--61\,294, 2018.

\bibitem{sunderhauf2016place}
N.~S{\"u}nderhauf, F.~Dayoub, S.~McMahon, B.~Talbot, R.~Schulz, P.~Corke,
  G.~Wyeth, B.~Upcroft, and M.~Milford, ``Place categorization and semantic
  mapping on a mobile robot,'' in \emph{ICRA}, 2016.

\bibitem{mancini2018robust}
M.~Mancini, S.~R. Bul{\`o}, B.~Caputo, and E.~Ricci, ``Robust place
  categorization with deep domain generalization,'' \emph{IEEE RA-L}, 2018.

\bibitem{zhou2017places}
B.~Zhou, A.~Lapedriza, A.~Khosla, A.~Oliva, and A.~Torralba, ``Places: A 10
  million {I}mage {D}atabase for {S}cene {R}ecognition,'' \emph{T. PAMI}, 2017.

\bibitem{redmon2018:arxiv}
J.~Redmon and A.~Farhadi, ``Yolov3: An incremental improvement,'' \emph{arXiv
  preprint arXiv:1804.02767}, 2018.

\bibitem{he2016deep}
K.~He, X.~Zhang, S.~Ren, and J.~Sun, ``Deep residual learning for image
  recognition,'' in \emph{CVPR}, 2016, pp. 770--778.

\bibitem{deng2009imagenet}
J.~Deng, W.~Dong, R.~Socher, L.-J. Li, K.~Li, and L.~Fei-Fei, ``Imagenet: A
  large-scale hierarchical image database,'' in \emph{CVPR}, 2009.

\bibitem{redmon2016you}
J.~Redmon, S.~Divvala, R.~Girshick, and A.~Farhadi, ``You only look once:
  Unified, real-time object detection,'' in \emph{CVPR}, 2016.

\bibitem{redmon2017yolo9000}
J.~Redmon and A.~Farhadi, ``Yolo9000: better, faster, stronger,'' in
  \emph{CVPR}, 2017.

\bibitem{liu2016ssd}
W.~Liu, D.~Anguelov, D.~Erhan, C.~Szegedy, S.~Reed, C.-Y. Fu, and A.~C. Berg,
  ``{SSD}: {S}ingle shot multibox detector,'' in \emph{ECCV}, 2016.

\bibitem{he2017mask}
K.~He, G.~Gkioxari, P.~Doll{\'a}r, and R.~Girshick, ``Mask {R-CNN},'' in
  \emph{ICCV}, 2017, pp. 2961--2969.

\bibitem{cai2018cascade}
Z.~Cai and N.~Vasconcelos, ``Cascade {R-CNN}: Delving into high quality object
  detection,'' in \emph{CVPR}, 2018, pp. 6154--6162.

\bibitem{lin2017focal}
T.-Y. Lin, P.~Goyal, R.~Girshick, K.~He, and P.~Doll{\'a}r, ``Focal loss for
  dense object detection,'' in \emph{ICCV}, 2017, pp. 2980--2988.

\bibitem{lin2014microsoft}
T.-Y. Lin, M.~Maire, S.~Belongie, J.~Hays, P.~Perona, D.~Ramanan,
  P.~Doll{\'a}r, and C.~L. Zitnick, ``Microsoft {COCO}: Common objects in
  context,'' in \emph{ECCV}, 2014.

\bibitem{zagoruyko2016wide}
S.~Zagoruyko and N.~Komodakis, ``Wide residual networks,'' \emph{arXiv preprint
  arXiv:1605.07146}, 2016.

\bibitem{xu2015show}
K.~Xu, J.~Ba, R.~Kiros, K.~Cho, A.~Courville, R.~Salakhudinov, R.~Zemel, and
  Y.~Bengio, ``Show, attend and tell: Neural image caption generation with
  visual attention,'' in \emph{ICML}, 2015, pp. 2048--2057.

\bibitem{song2015sun}
S.~Song, S.~P. Lichtenberg, and J.~Xiao, ``Sun {RGB-D}: A {RGB-D} scene
  understanding benchmark suite,'' in \emph{CVPR}, 2015, pp. 567--576.

\bibitem{silberman2012indoor}
N.~Silberman, D.~Hoiem, P.~Kohli, and R.~Fergus, ``Indoor segmentation and
  support inference from {RGBD} images,'' in \emph{ECCV}, 2012.

\bibitem{janoch2013category}
A.~Janoch, S.~Karayev, Y.~Jia, J.~T. Barron, M.~Fritz, K.~Saenko, and
  T.~Darrell, ``A category-level {3D} object dataset: Putting the kinect to
  work,'' in \emph{Consumer depth cameras for computer vision}, 2013.

\bibitem{xiao2013sun3d}
J.~Xiao, A.~Owens, and A.~Torralba, ``Sun3d: A database of big spaces
  reconstructed using {SFM} and object labels,'' in \emph{ICCV}, 2013.

\bibitem{krizhevsky2012imagenet}
A.~Krizhevsky, I.~Sutskever, and G.~E. Hinton, ``Imagenet classification with
  deep convolutional neural networks,'' in \emph{NIPS}, 2012.

\bibitem{fazl2012histogram}
E.~Fazl-Ersi and J.~K. Tsotsos, ``Histogram of oriented uniform patterns for
  robust place recognition and categorization,'' \emph{IJRR}, 2012.

\bibitem{yang2012object}
H.~Yang and J.~Wu, ``Object templates for visual place categorization,'' in
  \emph{ACCV}, 2012, pp. 470--483.

\bibitem{Trevor2014:ICRA}
A.~J. Trevor, J.~G. Rogers, and H.~I. Christensen, ``{Omni{M}apper: A modular
  multimodal mapping framework},'' in \emph{ICRA}.\hskip 1em plus 0.5em minus
  0.4em\relax IEEE, 2014.

\bibitem{neira2001data}
J.~Neira and J.~D. Tard{\'o}s, ``Data association in stochastic mapping using
  the joint compatibility test,'' \emph{IEEE Transactions on robotics and
  automation}, vol.~17, no.~6, pp. 890--897, 2001.

\bibitem{nieto2010:IROS}
C.~Nieto-Granda, J.~G. {Rogers III}, A.~J.~B. Trevor, and H.~I. Christensen,
  ``{Semantic Map Partitioning in Indoor Environments Using Regional
  Analysis},'' in \emph{IROS}.\hskip 1em plus 0.5em minus 0.4em\relax Taiwan:
  IEEE, October 2010.

\end{thebibliography}
